\documentclass[%
 aip,
rsi,%
 amsmath,amssymb,
 reprint,%
author-year,
groupedaddress
]{revtex4-1}

\usepackage{graphicx}
\usepackage{dcolumn}
\usepackage{bm}
\usepackage{capt-of}
\begin{document}
\title[Comparing Writing Styles using Word Embedding and Dynamic Time Warping]{Comparing Writing Styles using\\ Word Embedding and Dynamic Time Warping}

\author{Abhinav Tushar}
 \email[]{abhinav.tushar.vs@gmail.com}
\author{Abhinav Dahiya}
 \email[]{iit.abhinav.dahiya@gmail.com}
 \affiliation{
Department of Electrical Engineering\\
Indian Institute of Technology, Roorkee \\
(Both authors contributed equally)
}

\begin{abstract}
The development of plot or story in novels is reflected in the content and the words used. The flow of sentiments, which is one aspect of writing style, can be quantified by analyzing the flow of words. This study explores literary works as signals in word embedding space and tries to compare writing styles of popular classic novels using dynamic time warping.
\end{abstract}

\maketitle

\section{\label{intro}Introduction}
Writing style of a novel is based on many factors including the personal style of the author and the type of the novel itself. For example, {\sl thriller} novels by the same author tend to be similar in the way the story develops. Apart from the specific factual details, literary works tend to have a definite {\sl flow} of emotions that define the overall subjective {\sl feel} of the work as a whole.

This flow can be quantified and compared by analyzing the text using natural language processing techniques. This study uses word embedding models to generate time series for novels and then compare the resulting series using dynamic time warping to find similarities. Considering time series analysis rather than a pure statistical one can capture the {\sl flow} of the works and thus the similarities generated will provide a metric for comparing the novels based on the progression.

The following section explains the word embedding models and dynamic time warping. Section~\ref{approach} describes the approach followed. Results on few classic literary works are described in Section~\ref{res}. Conclusions are drawn in Section~\ref{conc}.

\section{\label{back}Background}
\subsection{Word Embedding}
Word embedding refers to techniques which allow representing words to vectors of real numbers in a continuous space. The vector representation can be learned using many techniques and relies on using the information based on the context, a word is present in.

We use a model proposed by \citet{Mikolov2013a,Mikolov2013b} that uses Continuous Bag-of-Words and Skip-Gram model to learn these continuous representations. This approach uses the following two models.

\subsubsection{Continuous Bag-of-Words model}
\begin{center}
  \includegraphics[width=0.3\textwidth]{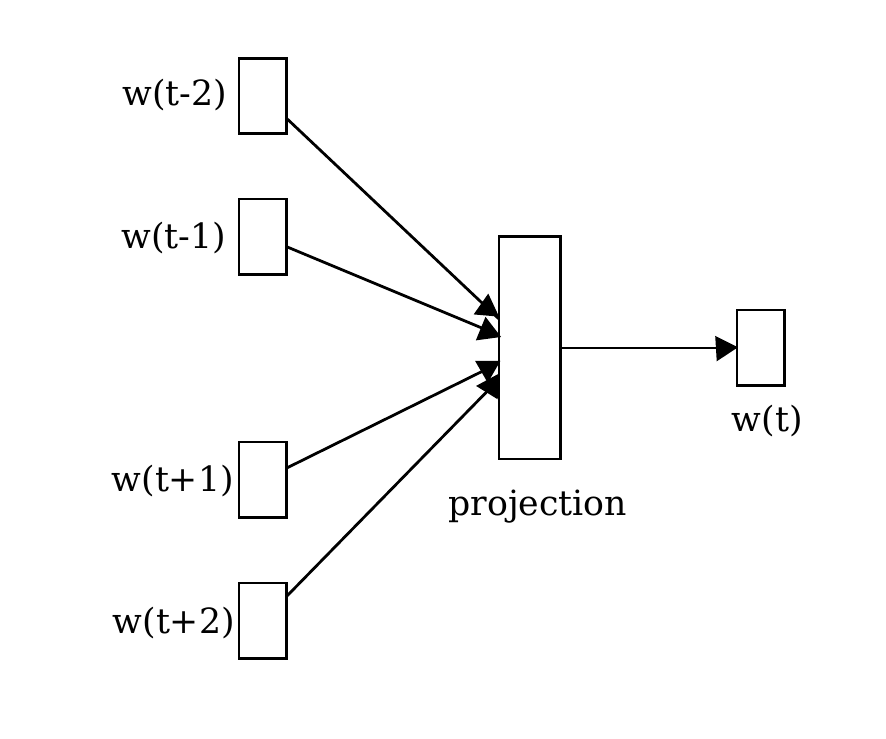}
  \captionof{figure}{Continuous Bag-of-Words model}
  \label{fig:cbow}
\end{center}
This model (figure~\ref{fig:cbow}) samples a context of size defined by $c$ from text around a word and tries to learn a projection to vector space that correctly predicts the middle word. Assuming words represented as one hot vectors $w(i)$, where $i$ denotes the position of the word in text, the model takes $w(t - c), w(t - (c - 1)), \cdots, w(t - 1), w(t + 1), \cdots, w(t + (c - 1)), w(t + c)$ and predicts the middle word $w(t)$.

\subsubsection{Skip-Gram model}
\begin{center}
  \includegraphics[width=0.3\textwidth]{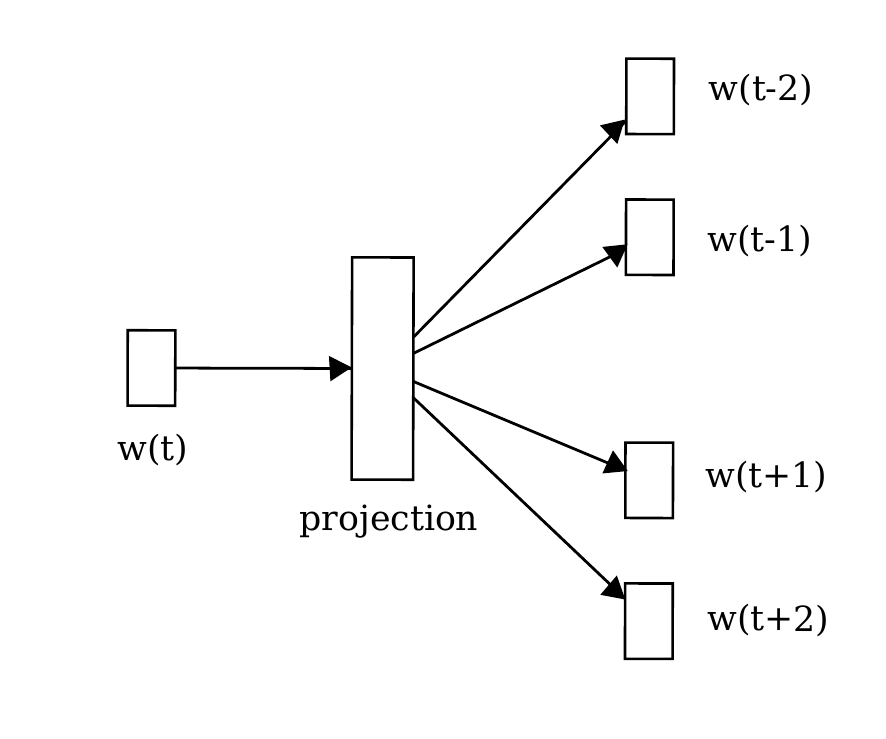}
  \captionof{figure}{Skip-Gram model}
  \label{fig:skip}
\end{center}
The skip-gram model (figure~\ref{fig:skip}) does the opposite of previous model. It tries to predict the context, given the middle word. Using the previous notation, this model takes $w(t)$ as input and outputs $w(t - c), w(t - (c - 1)), \cdots, w(t - 1), w(t + 1), \cdots, w(t + (c - 1)), w(t + c)$.

The projection layers are of desired vector dimension. The weights from one hot encoding of words to this layer provide, after training, the final transformation matrix for continuous word representation.

Once trained, words in the vector space are arranged according to semantic connections. This allows the vectors to have properties as shown below.

\begin{equation*}
W(king) - W(queen) \approx W(boy) - W(girl)
\end{equation*}

$W$ is the function from vocabulary to the vector space.

\subsection{Dynamic Time Warping}
{\sl Dynamic Time Warping} (DTW) is a technique to measure similarity between two time series. DTW handles the difference in speed and time between signals and has been used in applications like speech recognition and signature matching where the difference of signal speed should not affect the final result. It is based on optimal matching and the algorithm outputs a value corresponding to the separation of the two time series. One considerable advantage of DTW is that the two series need not to be of same length, which is the case here, as two different books will have different word count and consequently different length.

The objective of this algorithm is to calculate a distance measure for a given pair of temporal series, that can represent the similarity / dissimilarity between those two series.
This is done by determining a path $W$ which minimizes the cumulative euclidean distance between elements of the two series. Let the two temporal series, which need to be compared, are 

\begin{equation*}
A = a_1, a_2, a_3, ..., a_m \\
\end{equation*}
\begin{equation*}
B = b_1, b_2, b_3, ..., b_n
\end{equation*}

First, all the euclidean distances are calculated between each possible set of elements from the two series, which results in total of $m\times n$ values. Let the matrix depicting these distances be 

\begin{equation*}
D = 
 \begin{pmatrix}
  d_{1,1} & d_{1,2} & \cdots & d_{1,n} \\
  d_{2,1} & d_{2,2} & \cdots & d_{2,n} \\
  \vdots  & \vdots  & \ddots & \vdots  \\
  d_{m,1} & d_{m,2} & \cdots & d_{m,n} 
 \end{pmatrix}
\end{equation*}

where the element $d_{i,j}$ represents the euclidean distance between $a_i$ and $b_j$.
Now, using dynamic programming, the optimal path $W$ is determined from point $(1, 1)$ to $(m, n)$ along which the cumulative sum of the euclidean distances (i.e. the sum of $d_{i,j}$) is minimum. This path is continuous, which means that the indices of two consecutive elements of $W$ do not differ by more than one in either series. The path is determined using the following recursive function:

\begin{equation*}
\gamma(i,j) = d(i,j) + \min(\gamma(i-1,j-1), \gamma(i-1,j), \gamma(i,j-1))
\end{equation*}

where $\gamma(i,j)$ represents the cumulative sum up to elements $a_i$, and $b_j$. Figure~\ref{fig:dtw} shows an example using two sinusoidal series. The optimal path and the series are shown.

\begin{center}
  \includegraphics[width=0.5\textwidth]{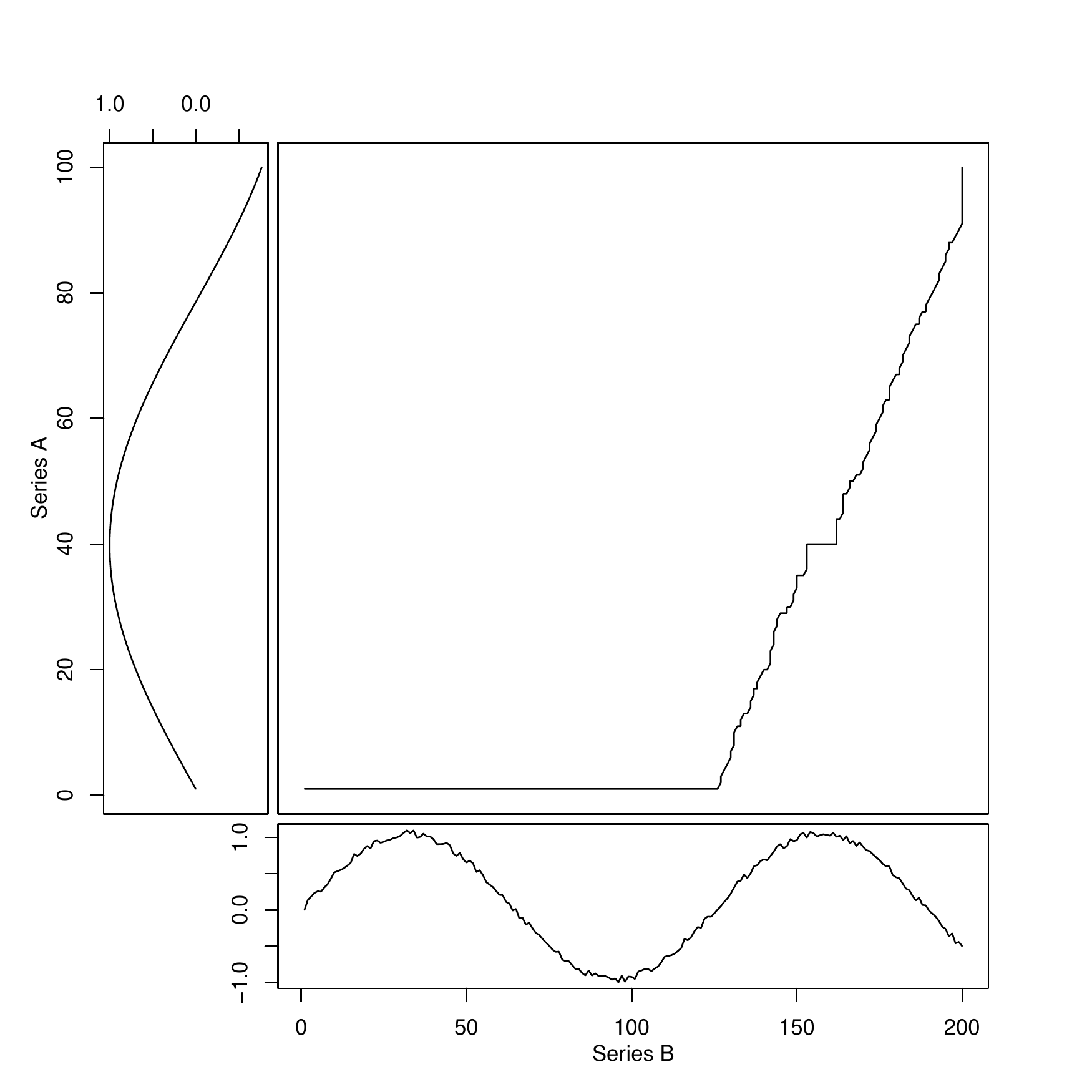}
  \captionof{figure}{Optimal path calculated using DTW for two sinusoidal series}
  \label{fig:dtw}
\end{center}

\section{\label{approach}Approach}
\subsection{Novels to Signal}

\begin{figure*}[!htb]
    \centering
    \includegraphics[width=\textwidth]{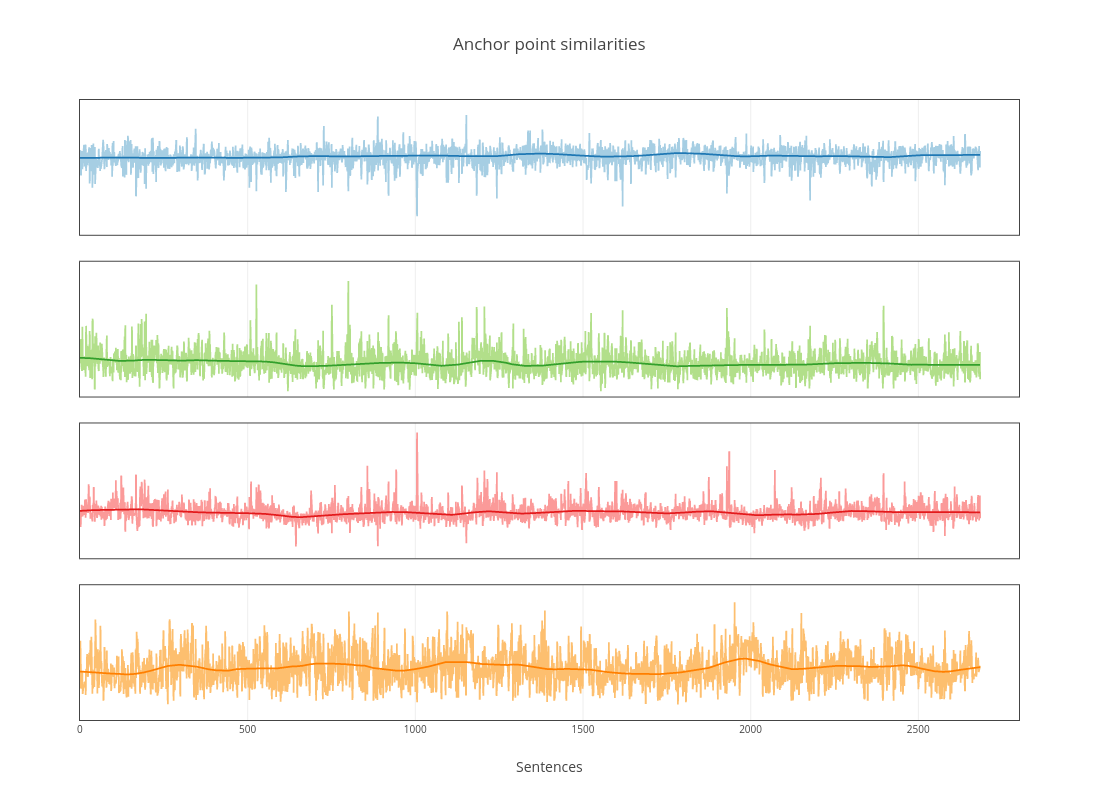}
    \caption{Anchor point similarities for {\sl The Sign of the Four}. Deeper color curves represent the filtered signal.}
    \label{fig:anchor}
\end{figure*}

Our approach for converting a novel to signal uses sentence vectors by taking mean of word vectors in a sentence. A word embedding model is trained using 1000 free e-books gathered from Project Gutenberg\footnote{Project Gutenberg \url{https://www.gutenberg.org}}. The model is trained using {\sl Word2Vec}\footnote{Python adaptation here \url{https://radimrehurek.com/gensim/models/word2vec.html}}.

Training is done using feature (word vector) size of 100 and context window size of 10 words. Once trained, a novel with $N_{w}$ words results in a 100 dimensional time series of length $N_{w}$.

To reduce the dimension of data word vectors in a sentence are averaged, resulting in a time series matrix of shape $N_{s} \times 100$, where $N_{s}$ is the number of sentences in the novel. To reduce the row dimensions, we find cosine similarity of each row with few {\sl anchor points} in the word space. This provides a measure of the movement of the signal in the whole vocabulary space.

For selecting appropriate anchor points to better cover the space in its entirety, we perform k-means clustering on the vocabulary of the embedding model and find 4 cluster centers. These 4 points act as the anchors, resulting in a final matrix of size $N_{s} \times 4$ for each book.

A plot of distances from these 4 points for {\sl The Sign of the Four} by {\sl Arthur Conan Doyle} is shown in figure~\ref{fig:anchor}. Due to the fickle nature of text a large amount of noise is present. After smoothing with a gaussian filter (window size = 200, $\sigma$ = 60), the essential rising and falling trend of curve is preserved. Filtered lines are shown with deeper color in the plot.

\subsection{Comparing Signals}
Once generated, the signals are compared using FastDTW \citep{Salvador2004} which implements a faster version of DTW algorithm. Vanilla DTW requires $O(n^{2})$ computation, while FastDTW computes in linear time.

We use 24 classic novels collected from Project Gutenberg (listed in table~\ref{table:books}) and generate distance matrix for the items.

\begin{table*}[!htb]
\centering
\begin{tabular}{l l r}
 {\bf S.no.} & {\bf Book Name} & {\bf Author} \\ \\
  1. & A Study in Scarlet & Arthur Conan Doyle \\
  2. & The Adventures of Sherlock Holmes & Arthur Conan Doyle \\
  3. & The Hound of the Baskervilles & Arthur Conan Doyle \\
  4. & The Return of Sherlock Holmes & Arthur Conan Doyle \\
  5. & The Sign of the Four & Arthur Conan Doyle \\
  6. & A Christmas Carol & Charles Dickens \\
  7. & A Tale of Two Cities & Charles Dickens \\
  8. & Bleak House & Charles Dickens \\
  9. & David Copperfield & Charles Dickens \\
  10. & Great Expectations & Charles Dickens \\
  11. & Hard Times & Charles Dickens \\
  12. & Oliver Twist & Charles Dickens \\
  13. & Emma & Jane Austen \\
  14. & Pride and Prejudice & Jane Austen \\
  15. & Persuasion & Jane Austen \\
  16. & Sense and Sensibility & Jane Austen \\
  17. & A Connecticut Yankee in King Arthur's Court & Mark Twain \\
  18. & Adventures of Huckleberry Finn & Mark Twain \\
  19. & Life on the Mississippi & Mark Twain \\
  20. & The Adventures of Tom Sawyer & Mark Twain \\
  21. & The Innocents Abroad & Mark Twain \\
  22. & The Mysterious Stranger & Mark Twain \\
  23. & The Prince and the Pauper & Mark Twain \\
  24. & Treasure Island & Robert Louis Stevenson \\
\end{tabular}
\caption{List of novels compared}
\label{table:books}
\end{table*}

\section{\label{res}Results}

\begin{center}
  \includegraphics[width=0.5\textwidth]{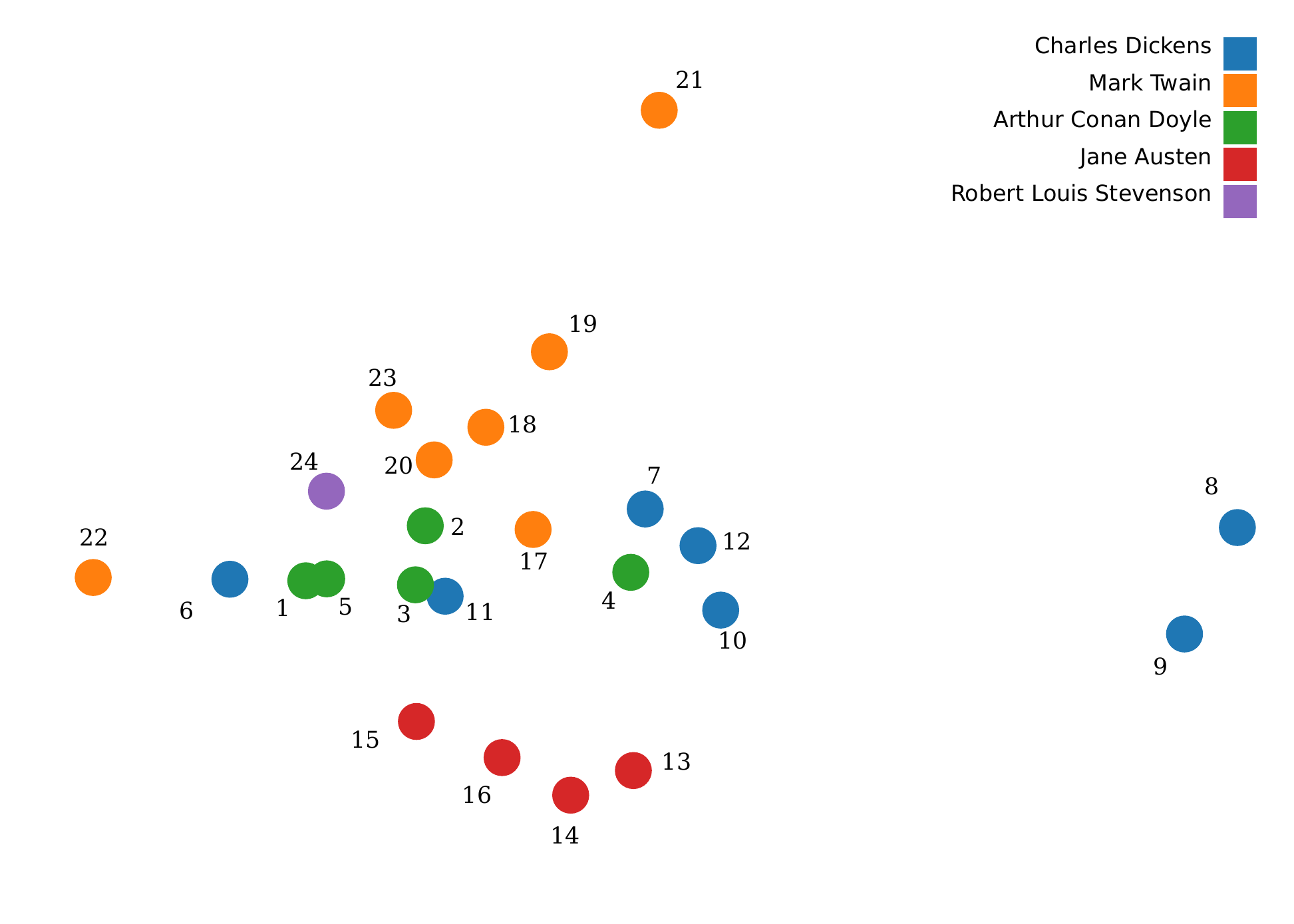}
  \captionof{figure}{Books scattered according to similarity. Look up table~\ref{table:books} for indices.}
  \label{fig:res}
\end{center}

A scatter plot of the books using multi dimensional scaling is shown in figure~\ref{fig:res}. Although the similarity measure itself is a useful metric for comparison, the scatter plot also clusters the authors according to the writing styles, affirming the hypothesis that a comparison of flow in the time series of word vectors can have subjective projections.

\section{\label{conc}Conclusions and Future Works}
This study provides a way to explore literary works as signals in word embedding space. The clusters of authors formed as the result of analysis provides encouraging support for this method as a high level text analysis technique.

A more rigorous study can be done by building on present method to identify the principle components involved in shaping the overall picture of a book while at the same time, being agnostic of factual details. The text representation can be made better using paragraph (or sentence) vectors instead of word vectors. Anchor points can be improved based on the effects and the number of points. A frequency based analysis of signals can provide a better insight.

\section*{References}

\bibliography{aipsamp}

\end{document}